\documentclass[10pt,twocolumn,letterpaper]{article}

\usepackage{wacv}              
\usepackage[accsupp]{axessibility}  

\usepackage{graphicx}
\usepackage{amsmath}
\usepackage{amssymb}
\usepackage{booktabs}

\usepackage{multirow}
\usepackage{hhline}
\usepackage{arydshln}
\usepackage{comment}
\usepackage{enumitem}

\usepackage[capitalize]{cleveref}
\crefname{section}{Sec.}{Secs.}
\Crefname{section}{Section}{Sections}
\Crefname{table}{Table}{Tables}
\crefname{table}{Tab.}{Tabs.}

\begin{document}

\title{Arbitrary-Resolution and Arbitrary-Scale Face Super-Resolution with Implicit Representation Networks}

\author{Yi Ting Tsai, Yu Wei Chen, Hong-Han Shuai, and Ching-Chun Huang\\
National Yang Ming Chiao Tung University\\
{\tt\small \{tsai.cs09, agarya89.11, chingchun\}@nycu.edu.tw}\quad
{\tt\small hhshuai@nctu.edu.tw}
}
\maketitle

\begin{abstract}
Face super-resolution (FSR) is a critical technique for enhancing low-resolution facial images and has significant implications for face-related tasks. However, existing FSR methods are limited by fixed up-sampling scales and sensitivity to input size variations. To address these limitations, this paper introduces an Arbitrary-Resolution and Arbitrary-Scale FSR method with implicit representation networks (ARASFSR), featuring three novel designs. First, ARASFSR employs 2D deep features, local relative coordinates, and up-sampling scale ratios to predict RGB values for each target pixel, allowing super-resolution at any up-sampling scale. Second, a local frequency estimation module captures high-frequency facial texture information to reduce the spectral bias effect. Lastly, a global coordinate modulation module guides FSR to leverage prior facial structure knowledge and achieve resolution adaptation effectively. Quantitative and qualitative evaluations demonstrate the robustness of ARASFSR over existing state-of-the-art methods while super-resolving facial images across various input sizes and up-sampling scales.
\end{abstract}

\section{Introduction}

\begin{figure*}[h!]
  \centering
  \includegraphics[width=\linewidth]{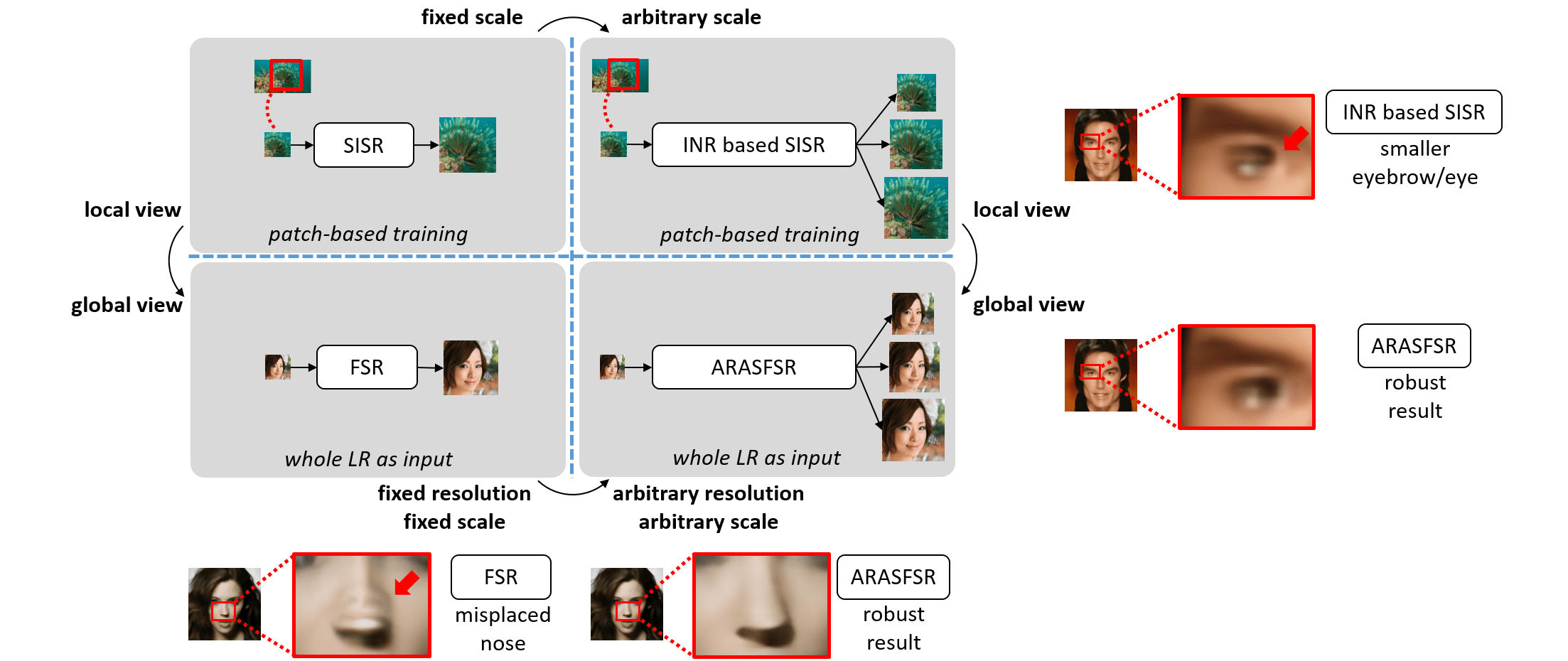}
  \caption{Comparison of our proposed arbitrary-resolution and arbitrary-scale face super-resolution (ARASFSR) method with single image super-resolution (SISR) methods, face super-resolution (FSR) methods, and implicit neural representation (INR) based SISR methods. The figure highlights the differences in patch-based training in SISR versus the need for the whole low-resolution face image as input to match the test-time input distribution in FSR, resulting in a global approach instead of SISR's local view. Conventional FSR methods have limited practical applications due to their fixed up-sampling scales and sensitivity to input resolution variations, while applying INR-based SISR methods to face images is not straightforward due to the lack of a global view. To address these issues, we propose ARASFSR, which utilizes implicit representation networks to produce facial images of any resolution and can handle changes in input resolution.} 
  \label{fig:teaser}
\end{figure*}

Face super-resolution (FSR) is a technique that focuses on enhancing the resolution and quality of low-resolution facial images. This process involves generating high-resolution (HR) facial images from their low-resolution (LR) counterparts, offering significant advantages in various applications, such as face detection, face recognition, and face parsing. Due to its potential applications, FSR has garnered substantial attention in recent years. 

FSR is a specific type of single image super-resolution (SISR) problem. As illustrated in Fig.~\ref{fig:teaser}, when addressing the SISR problem, it is assumed that the dataset includes general natural scenes and is not specific to any particular category. Consequently, state-of-the-art (SOTA) SISR methods focus on understanding local image features and learning the super-resolution mapping from LR patches to HR patches. Therefore, patch-based SISR is essentially suitable for handling LR images with varying resolutions. However, facial images possess a more consistent global structure. To leverage this prior knowledge, a more effective FSR method may prefer taking the entire and aligned LR facial image as input and learning the mapping globally to its HR counterpart. However, this global property makes the conventional FSR method sensitive to input image resolution. As a result, the design of FSR necessitates a different approach compared to SISR.

In recent years, deep learning-based methods have made significant strides in improving the performance of FSR. Attention mechanisms have been extensively employed in various FSR methods \cite{ChenSPARNet, lu2021face, MRRNet} to enhance facial structural information and texture details. Moreover, some FSR methods leverage face priors \cite{CT-FSRNet-2018, progressive-face-sr, ma2020deep}, such as landmarks and component heatmaps, to further refine the recovery quality. Despite these advances, existing FSR methods exhibit limitations, such as their applicability only to a specific up-sampling scale, which constrains their generalizability. Furthermore, these FSR methods process the entire LR face image as input rather than in patches, making them highly sensitive to variations in input resolution. Consequently, as demonstrated in Fig.~\ref{fig:teaser}, changes in input resolution can considerably impact the performance of FSR methods, making them impractical. This implies that when there are changes in up-sampling scales or input resolutions, the existing FSR methods require the network to be retrained.

Facial images in the real world come in various resolutions, presenting challenges for downstream facial-related tasks. To address this issue, it is crucial to develop flexible FSR methods capable of handling a wider range of up-sampling scales and input sizes. For instance, in low-resolution face recognition tasks, some methods \cite{superIdentity, IdentityPreserved, IdentityAware, bao2022distilling} first super-resolve LR facial images into HR ones and then perform feature extraction. However, existing FSR methods are limited to a fixed up-sampling scale, while feature extraction networks typically have a fixed input size. As a result, the HR output of FSR must be resized before being fed into the feature extraction network, which compromises data fidelity and may negatively impact the downstream task. Therefore, there is an urgent need to develop a generalized FSR method that can handle a broader range of up-sampling scales and input sizes, making FSR more suitable for various facial-related applications.

Implicit neural representation (INR) has gained popularity due to its effectiveness in 3D shape reconstruction, as demonstrated in recent works such as \cite{sitzmann2019scene, mildenhall2020nerf, Local_Implicit_Grid_CVPR20, chibane20ifnet, Genova_2020_CVPR}. The fundamental concept behind INR involves representing objects as multi-layer perceptrons (MLPs) that map coordinates to corresponding signal values. In recent years, there has been growing interest in using INR to learn 2D continuous image representations \cite{DBLP:journals/corr/abs-2012-09161, xu2021ultrasr, 9776607, lte-jaewon-lee, nguyen2023single}. This approach utilizes implicit image functions, which are MLPs that input local coordinate information and query local latent codes to predict RGB values at target coordinates. This coordinates continuity property enables the representation of images in arbitrary resolutions, making it possible to achieve super-resolution at any up-sampling scale. However, existing methods based on local implicit image functions are not directly applicable to FSR due to their lack of a global view and designs that incorporate facial prior knowledge. Furthermore, as illustrated in Fig.~\ref{fig:teaser}, these methods are sensitive to input image resolution, similar to existing FSR methods.

In this paper, we propose an Arbitrary-Resolution and Arbitrary-Scale FSR method with implicit representation networks (ARASFSR) to address the aforementioned issues. Our method takes 2D deep features, local relative coordinates, and up-sampling scale ratios as inputs of the implicit representation network to predict the RGB value for each target pixel, enabling super-resolution at any up-sampling scale. To further enhance FSR reconstruction, we designed two specialized modules. The first is a local frequency estimation module that predicts high-frequency information about facial texture to reduce the spectral bias effect. The second is a global coordinate modulation module that guides facial structure, allowing for leveraging prior facial knowledge and achieving input resolution adaptation. We conduct comprehensive quantitative and qualitative evaluations, demonstrating our proposed method's robustness and superiority over SOTA methods. Our main contributions can be summarized as follows:
\begin{itemize}[leftmargin=10pt, parsep=-3pt]
\item We propose ARASFSR, a novel FSR method, with implicit representation networks to address the limitations of FSR in terms of fixed up-sampling scales and sensitivity to input resolution variations. 
\item The method incorporates two specialized modules to learn high-frequency information and guide facial structure for arbitrary up-sampling scales and variations in input resolutions, resulting in detailed and realistic facial image reconstruction. 
\item The proposed method outperforms state-of-the-art methods in comprehensive quantitative and qualitative evaluations, demonstrating its robustness and superiority in generalized FSR tasks. 
\end{itemize}

\section{Related Works}
\subsection{Face Super-Resolution}
Deep learning-based methods have made significant strides in the field of FSR. URDGN \cite{yu2016ultra} uses a deep discriminative generative network to super-resolve ultra-LR facial images. Wavelet-SRNet \cite{huang2017wavelet} employs a wavelet-based CNN method to predict the corresponding series of HR wavelet coefficients from an LR input image. Some methods incorporate facial priors in the super-resolution process. FSRNet \cite{CT-FSRNet-2018} utilizes facial landmark heatmaps and parsing maps. PFSRNet \cite{progressive-face-sr} proposes a facial attention loss by multiplying the pixel difference and heatmap values to progressively recover facial details. DIC \cite{ma2020deep} adopts an iterative collaboration based on the guidance of landmark maps. Attention mechanisms have also been adopted in the latest methods. SPARNet \cite{ChenSPARNet} makes use of a spatial attention mechanism to adaptively guide features associated with facial structures. SISN \cite{lu2021face} introduces a split-attention in the split-attention network to strengthen facial structure and details. MRRNet \cite{MRRNet} designs a spatial attention mechanism guided by multi-scale receptive field features. 


\subsection{Implicit Neural Representation (INR)}
INR is an idea that represents an object using a function that maps coordinates to their corresponding signal values. The success of INR in 3D tasks has led to increasing interest in using INR for 2D image representations. IMNET \cite{chen2018implicit_decoder} introduces an implicit field decoder for 2D shape generation. SIREN \cite{sitzmann2020implicit} leverages sine instead of a ReLU as activation functions for implicit neural representations, resulting in higher image fidelity. LIIF \cite{DBLP:journals/corr/abs-2012-09161} proposes a local implicit image function to learn continuous image representation by taking coordinate and local latent codes as input to predict RGB values at target coordinates, achieving super-resolution at arbitrary up-sampling scale. LTE \cite{lte-jaewon-lee} designs a dominant-frequency estimator for implicit image function to alleviate spectral bias \cite{rahaman2019spectral} of a standard MLP with ReLU. Finally, DIINN \cite{nguyen2023single} decouples content and positional features using dual interactive implicit neural networks.

However, applying these methods \cite{DBLP:journals/corr/abs-2012-09161, lte-jaewon-lee, nguyen2023single} to face images is not straightforward. While coordinates are used to enhance features in LTE \cite{lte-jaewon-lee}, they are not used directly as input to MLPs, making it less robust when the up-sampling scale changes. Although LIIF \cite{DBLP:journals/corr/abs-2012-09161} and DIINN \cite{nguyen2023single} take local relative coordinates as MLP's input, the local implicit image functions still lack a global view, leading to artifacts in the output when the input image size changes. Inspired by the effectiveness of the implicit image function in SISR, ARASFSR extends the flexibility of continuous image representation to face super-resolution. Different from the aforementioned works, ARASFSR has a specific design tailored to face images to alleviate the problem of change in up-sampling scales and be more robust to input size variations.

\section{Method}

\subsection{Overview}
\begin{figure*}[h!]
  \centering
  \includegraphics[width=\textwidth]{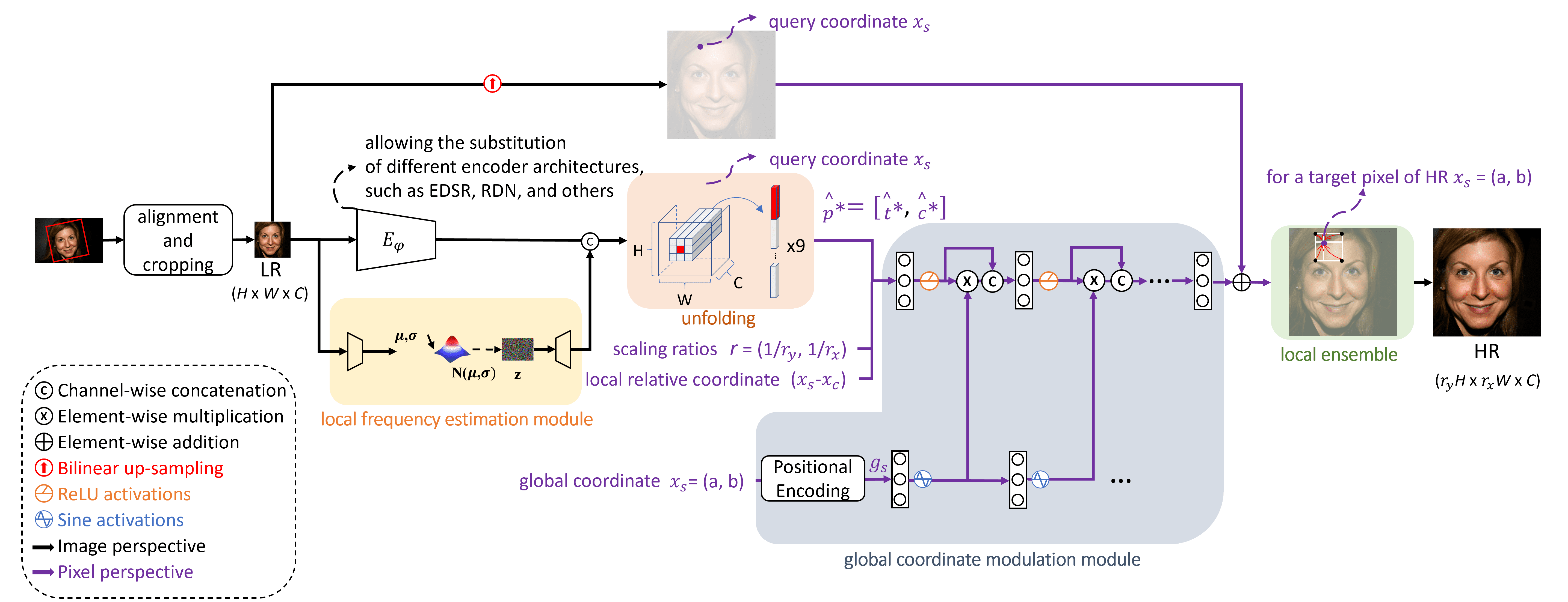}
  \caption{Overall architecture of the proposed framework.}
  \label{fig:framework}
\end{figure*}

Our goal is to develop a general framework for FSR that generates high-quality facial images at arbitrary up-sampling scales while accommodating variations in input resolution. Fig.~\ref{fig:framework} illustrates the overall architecture of the proposed framework. To achieve arbitrary-scale FSR, we introduce ARASFSR, which utilizes an implicit image function to represent facial images with arbitrary resolution. Besides the modified implicit representation network, our framework integrates two crucial modules: (a) a local frequency estimation module, which captures high-frequency information for facial texture to reduce the spectral bias effect; (b) a global coordinate modulation module, which allows for leveraging prior facial structure knowledge and enabling operation across various input sizes.

\subsection{Implicit Image Function}
\label{sec:Implicit Image Function}
In Local Implicit Image Function approaches \cite{DBLP:journals/corr/abs-2012-09161, lte-jaewon-lee, nguyen2023single}, the decoding function $f_{\theta}$ is shared among all images and is parameterized using an MLP with $\theta$ as its model parameters. It maps the latent feature codes and local coordinates to RGB values, i.e., $f_{\theta }(c,x): (\mathcal{C},\mathcal{X})\mapsto \mathcal{S}$, where $c \in \mathcal{C}$ is a latent feature code from an encoder $E_{\varphi }$ with $\varphi$ as its parameters, $x \in \mathcal{X}$ is a 2D coordinate in the continuous image domain, and $\mathcal{S}$ is a space of predicted RGB values from $f_{\theta}$.

Our method aims to super-resolve LR facial images $I^{LR} \in \mathbb{R}^{H\times W \times C}$ to HR images $I^{HR} \in \mathbb{R}^{r_yH\times r_xW \times C}$ at any fractional ratio $r_{y}$ and $r_{x}$ and arbitrary input resolution $H$ and $W$. Based on the conventional Local Implicit Image Function approaches with modification, for a continuous image $I^{(i)}$, the RGB value $s$ at the target coordinate $x_{s}$ is defined as follows:
\begin{equation}
\label{eq:LIIF}
s = I^{(i)}(x_{s}) = f_{\theta}(c^{\ast }, x_{s} - x_{c}),
\end{equation}
where $c^{\ast }$ represents the nearest (Euclidean distance) latent feature code from $x_{s}$, and $x_{c}$ denotes the coordinate of the latent code $c^{\ast }$ in the image domain. 

\noindent\textbf{Feature unfolding.} To enrich the feature map, we apply feature unfolding \cite{DBLP:journals/corr/abs-2012-09161, nguyen2023single}, which involves concatenating the features from a 3 $\times$ 3 neighborhood around each pixel of the feature map. Accordingly, Equation (\ref{eq:LIIF}) is replaced by:
\begin{equation}
\label{eq:LIIF with unfolding}
s = I^{(i)}(x_{s}) = f_{\theta}(\hat{c}^{\ast }, x_{s} - x_{c}),
\end{equation}
where $\hat{c}^{\ast }$ is the enriched local feature code after increasing the number of channels.

\noindent\textbf{Scaling ratio.} Although $f_{\theta }(c,x)$ can effectively model the continuous distribution, spectral bias tends to make $f_{\theta}$ learn a smoother function (i.e., low-frequency function), thereby losing its capacity to recover high-frequency details. To tackle this issue, we propose incorporating an additional scaling ratio as a conditional input to the decoding function. This conditional factor enhances the representational capacity of the original $f_{\theta }$. When the scaling ratio is small, $f_{\theta }$ is inclined to model smoother functions (less high frequency); conversely, $f_{\theta }$ is encouraged to model functions with more high frequency when the scaling ratio is large. Consequently, Equation~(\ref{eq:LIIF with unfolding}) is extended to:
\begin{equation}
\label{eq:LIIF with scaling ratio}
s = I^{(i)}(x_{s}) = f_{\theta}(\hat{c}^{\ast }, x_{s} - x_{c}, r),
\end{equation}
where $r = \left [ 1/r_{y}, 1/r_{x} \right ]$ represents the scaling ratio.

\noindent\textbf{Local ensemble.} The signal prediction at $x_{s}$ is done by querying the nearest latent code, and as $x_{s}$ moves in the 2D domain, the selection of the nearest latent code can suddenly change, causing the issue of discontinuous prediction. To address this, we refer to \cite{DBLP:journals/corr/abs-2012-09161, lte-jaewon-lee} and use local ensemble. Accordingly, we modify Equation~(\ref{eq:LIIF with scaling ratio}) as follows:
\begin{equation}
\label{eq:LIIF with local ensemble}
s = I^{(i)}(x_{s}) = \sum_{k \in K} w_k f_{\theta}(\hat{c}^{\ast }_k, x_{s} - x_{c_k}, r),
\end{equation}
where $K$ is a set of indices of the four nearest latent codes around $x_{s}$, $w_K$ is the bilinear interpolation weight for the enriched latent feature code $\hat{c}_k$, such that $\sum_k w_k = 1$.

\subsection{Local Frequency Estimation Module}
\label{sec:Local Frequency Estimation Module}
As previously mentioned, spectral bias is a challenge where a standalone MLP struggles to capture high-frequency textures. To overcome this limitation, we propose a local frequency estimation module. This module employs an encoder-decoder architecture to predict the conditional distribution of high-frequency details ($t$) given an LR facial image $I^{LR}$, specifically, $P(t|I^{LR})$. In particular, the encoder is designed to predict the parameters of a multi-dimensional Gaussian distribution based on the content of $I^{LR}$, which models the conditional distribution of the frequency latent code $P(z|I^{LR})$. Consequently, for different $I^{LR}$ instances, the encoder can estimate image-specific high-frequency latent code $z$ in a probabilistic manner. By sampling a latent code $z$ from $P(z|I^{LR})$ and inputting it into the decoder, we obtain a frequency token $t$.

After performing feature unfolding, we denote the estimated image-specific high-frequency token of a reference pixel location as $\hat{t}^{\ast}$. By concatenating $\hat{t}^{\ast}$ with the latent feature code $\hat{c}^{\ast}$, we enhance the representation function $f_{\theta}$ and update Equation (\ref{eq:LIIF with local ensemble}) as follows:
\begin{equation}
\label{eq:LIIF with local module}
s = I^{(i)}(x_{s}) = \sum_{k \in K} w_k f_{\theta}(\hat{p}^{\ast }_k, x_{s} - x_{c_k}, r),
\end{equation}
where $\hat{p}^{\ast } = \left [ \hat{t}^{\ast }, \hat{c}^{\ast } \right ]$ represents the high-frequency enhanced content feature.

\subsection{Global Coordinate Modulation Module}
\label{sec:Global Coordinate Modulation Module}
SISR is typically performed using patch-based training, wherein small patches of LR images are used to generate corresponding HR patches. However, in FSR, the entire LR facial image must be mapped to an HR counterpart. Previous SR methods based on INR \cite{DBLP:journals/corr/abs-2012-09161, lte-jaewon-lee, nguyen2023single} suffer from artifacts when the input resolution of facial images changes because the implicit image function only possesses a local view. In contrast, FSR methods process the entire LR facial image as input, making the global coordinate an essential component that provides landmark locations for reconstructing facial details. Furthermore, the global coordinate is resilient to variations in up-sampling scales and input sizes, making it an ideal source of facial priors.

Intuitively, if facial images are pre-aligned, a well-trained model may be able to predict the possible texture around a specific location based on the normalized coordinate and facial structure. To provide the implicit image function with a global perspective, we propose a global coordinate guidance module that incorporates the global coordinate $x_s$ as input. In our implementation, we employ positional encoding \cite{mildenhall2020nerf, xu2021ultrasr}, as illustrated in Equation (\ref{eq:positional_encoding}), to map the coordinates to a higher dimension before inputting them into the MLP. Note that each of the two coordinate values in $x_s$ is normalized to lie within the range $\left[-1, 1\right]$ before positional encoding. In our experiments, we set $N = 10$.
\begin{equation}
\begin{split}
\label{eq:positional_encoding}
g_s = (x_s, sin(2^0\pi x_s), cos(2^0\pi x_s),...,\\
sin(2^{N-1}\pi x_s), cos(2^{N-1} \pi x_s)).
\end{split}
\end{equation}

Recent approaches have addressed the spectral bias issue by employing non-linear activations. SIREN \cite{sitzmann2020implicit} utilizes the sine layer, which leads to rapid convergence and high data fidelity. Additionally, works such as \cite{mehta2021modulated, nguyen2023single} adopted modulated periodic activations. Inspired by them, we implement periodic activations to predict the modulation parameters given $g_s$. Next, as illustrated in Fig.~\ref{fig:framework}, we modulate the local implicit function $f_{\theta}$ by multiplying and concatenating in order to fuse the original feature of the MLP with the encoded global coordinate information. The process is detailed in the following equations:


\begin{equation}
\label{eq:s0}
s_0 = ReLU(w_0 \left [ \hat{p}^{\ast }_k, x_{s} - x_{c_k}, r \right ]+b_0).
\end{equation}
\begin{equation}
\label{eq:g0}
\hat{g}_0 = sin(w_0' g_s + b_0').
\end{equation}
\begin{equation}
\label{eq:m0}
m_0 = s_0 \odot \hat{g}_0.
\end{equation}
\begin{equation}
\label{eq:si}
s_i = ReLU(w_i\left [ m_{i-1}, s_{i-1} \right ]+b_i).
\end{equation}
\begin{equation}
\label{eq:gi}
\hat{g}_i = sin(w_i' g_s + b_i').
\end{equation}
\begin{equation}
\label{eq:mi}
m_i = s_i \odot \hat{g}_i.
\end{equation}

In the above equations, $w_i$ and $w_i'$ are the weights, $b_i$ and $b_i'$ are the biases, and $s$ is the latent feature of the $i^{th}$ layer within the MLP. $\hat{g}_i$ is the global positional feature and $m_i$ is the modulated output. The last output of $s_i$ is then passed through a final dense layer to output the predicted RGB value. With the guidance of the global coordinate, the modulated implicit image function now has a global view to identify the potential location of the landmark.

In summary, the global coordinate modulation module incorporates prior facial structure information into FSR by mapping the global coordinate into a high-dimensional space using position encoding. By combining global coordinate guidance with locally enhanced features, our ARASFSR achieves improved facial reconstruction results.

\subsection{Skip Connection}
The effectiveness of long skip connections in learning high-frequency components and improving convergence stability has been demonstrated in residual networks \cite{kim2016accurate,lte-jaewon-lee}. In our proposed architecture, we introduce an additional branch incorporating a skip connection. Including this long skip connection can alleviate information loss and enhance the network's capability to capture fine details.

\subsection{Loss Function}
We train our model end-to-end and use the Charbonnier loss \cite{lai2017deep} given by:
\begin{equation}
\label{eq:Charbonnier loss}
L_{\delta} = \frac{1}{N}\sum_{i=1}^{N}\sqrt{(I^{HR}(x_{s}) - I^{SR}(x_{s}))^2 + \delta^2},
\end{equation}
where $I^{HR}(x_{s})$ is the ground truth value, $I^{SR}(x_{s})$ is the predicted value, $N$ is the total number of samples, and $\delta$ is a hyper-parameter that controls the smoothness of the loss function.

\section{Experiments}
\subsection{Datasets and Metrics}
We carry out experiments on four datasets. Firstly, we compare the performance of our approach with INR-based SISR methods at different up-sampling scales on CelebAHQ dataset \cite{karras2017progressive}. This dataset consists of 30,000 HR face images (1024×1024 pixels) selected from CelebA dataset \cite{liu2015deep}. Secondly, we create CelebAHQ-NN-JPEG dataset by down-sampling CelebAHQ dataset using the nearest neighbor method and introducing JPEG compression artifacts. This dataset, along with SCface dataset \cite{grgic2011scface} containing facial images captured by surveillance cameras, demonstrates our method's applicability in real-world scenarios. Lastly, we compare our approach with FSR methods on CelebAHQ and Helen datasets \cite{le2012interactive}. We evaluate the performance using two widely-used metrics, PSNR and SSIM \cite{wang2004image}, calculated on the Y channel in the YCbCr color space.

\subsection{Implementation Details}
\begin{table}[h!]\small
\centering
\caption{Quantitative comparison on CelebAHQ with INR-based SISR methods (PSNR(dB)). The best and second best performances are highlighted in \textcolor{red}{red} and \textcolor{blue}{blue} colors, respectively. "E-" and "R-" indicate the use of EDSR and RDN as encoders, respectively.}
\label{tab:ideal case on CelebAHQ}
\begin{tabular}{c|cc|cc}
                  & \multicolumn{2}{c|}{In-distribution} & \multicolumn{2}{c}{Out-of-distribution} \\
Method            & ×1.5          & ×2            & ×4        & ×8        \\
                  & 64-96         & 64-128        & 64-256    & 64-512    \\ \hhline{=====}
E-LIIF \cite{DBLP:journals/corr/abs-2012-09161}         & 40.1002       & 36.9650       & \textcolor{blue}{32.3501}   & \textcolor{blue}{30.8161}   \\
E-LTE \cite{lte-jaewon-lee}         & \textcolor{blue}{40.1913}       & \textcolor{blue}{37.1062}       & 31.2708   & 28.4518   \\
E-DIINN \cite{nguyen2023single}       & 40.1781       & 37.0943       & 32.1410   & 30.4560   \\
E-ARASFSR & \textcolor{red}{40.1954}       & \textcolor{red}{37.1097}       & \textcolor{red}{32.4799}   & \textcolor{red}{30.9148}   \\ \hline
R-LIIF \cite{DBLP:journals/corr/abs-2012-09161}         & 40.2433       & 37.1188       & \textcolor{blue}{32.3950}   & \textcolor{blue}{30.7763}   \\
R-LTE \cite{lte-jaewon-lee}          & 40.2813       & \textcolor{red}{37.2104}       & 31.9483   & 29.9873   \\
R-DIINN \cite{nguyen2023single}         & \textcolor{blue}{40.3033}       & 37.1676       & 32.0622   & 30.6020   \\
R-ARASFSR & \textcolor{red}{40.3100}       & \textcolor{blue}{37.1830}       & \textcolor{red}{32.4987}   & \textcolor{red}{30.9741}  
\end{tabular}
\end{table}
Since CelebAHQ is a high-quality dataset with pre-aligned faces, no further alignment is necessary. For Helen and SCface, we align and crop all facial images with respect to their landmarks using MTCNN \cite{7553523}. In FSR, an LR face image of size $L_{r}$×$L_{r}$ is mapped to its HR counterpart of size $H_{r}$×$H_{r}$. To evaluate the effectiveness of different up-sampling scales, multiple LR-HR pairs are required. To achieve this, the ground-truth images ($D^{GT}$) (e.g., 1024×1024 high-quality images in CelebAHQ) are down-sampled using bicubic interpolation to produce the corresponding HR images ($D^{HR}$) at variant target HR levels $\{ H_{r}\}$. Similarly, we attain LR images ($D^{LR}$) at different target LR levels $\{L_{r}\}$ by using either bicubic or nearest-neighbor down-sampling with augmented JPEG compression noise. However, to validate model generalization across input resolution, we fix the target $L_{r}$ level and vary the target $H_{r}$ level according to up-sampling scales during training. Furthermore, we use Charbonnier loss \cite{lai2017deep} and Adam optimizer \cite{kingma2014adam}. The models are trained for 200 epochs with a batch size of 16.  The initial learning rate is 1e-4 and is reduced by a factor of 0.1 at epoch 100.

\subsection{Comparison with State-of-the-Arts}

\subsubsection{Comparison with INR-based SISR methods}
\label{sec:Comparison with INR-based SISR methods}
We compare our proposed method with SOTA INR-based SISR methods such as LIIF \cite{DBLP:journals/corr/abs-2012-09161}, LTE \cite{lte-jaewon-lee}, and DIINN \cite{nguyen2023single}, on different scenarios.
\begin{figure}[h!]
  \centering
  \includegraphics[width=\linewidth]{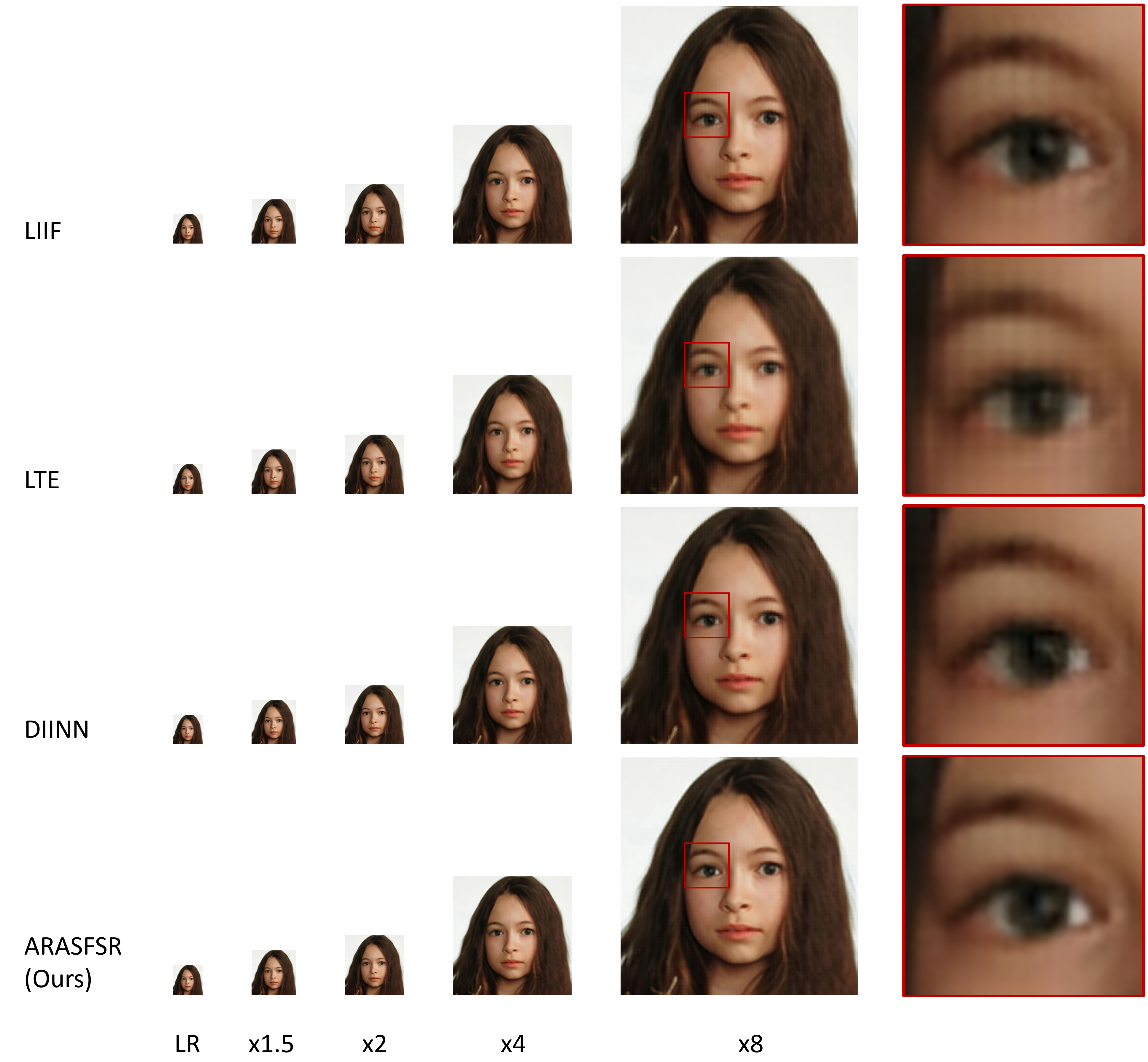}
  \caption{Visual comparison on CelebAHQ with INR-based SISR methods.}
  \label{fig:ideal case on CelebAHQ}
\end{figure}

\begin{figure}[h!]
  \centering
  \includegraphics[width=\linewidth]{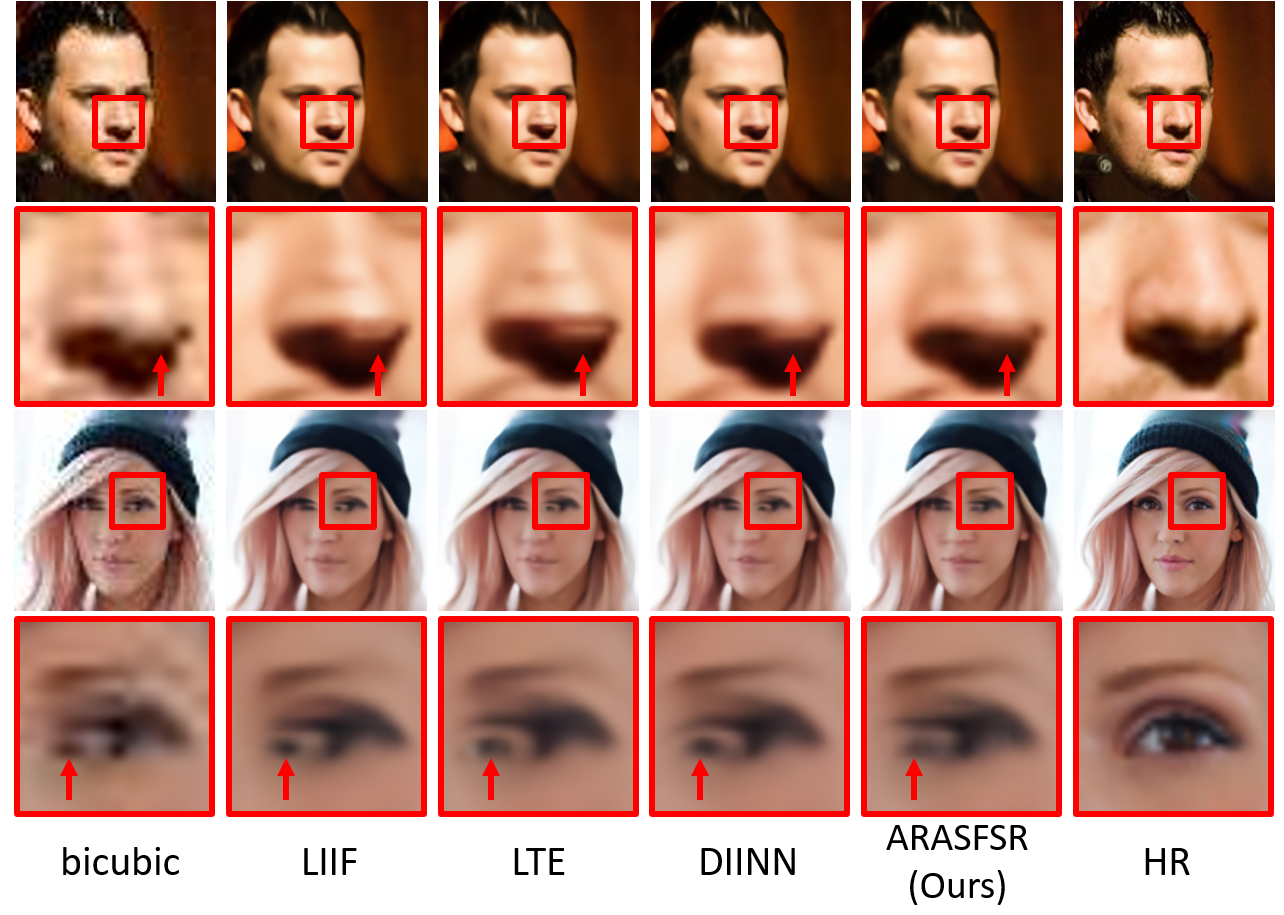}
  \caption{Visual comparison of real world case on CelebAHQ-NN-JPEG with INR-based SISR methods.}
  \label{fig:real world case on CelebAHQ-NN-JPEG}
\end{figure}

\noindent\textbf{Evaluation on CelebAHQ:} To evaluate the performance of \emph{various up-sampling scales} with INR-based SISR methods, we assess up-sampling tasks involving both in-distribution and out-of-distribution scales. Considering that the average image resolution of large face datasets, such as Vggface2 \cite{Vggface2}, is typically below 180×180 pixels without background cropping, we uniformly sample up-sampling scales within $\{\times1\sim\times2\}$ and set $L_{r}=64$ and $H_{r}\in\{64\sim128\}$ during training. During testing, we evaluate our method with much larger up-sampling scales within $\{\times1.5\sim\times8\}$. Table \ref{tab:ideal case on CelebAHQ} presents a quantitative comparison on CelebAHQ. The top and bottom rows show the results when EDSR \cite{lim2017enhanced} and RDN \cite{zhang2018residual} are used as encoders, respectively. ARASFSR achieves comparable performance for in-distribution scales and outperforms other methods for out-of-distribution scales. This demonstrates the superior generalizability of ARASFSR to arbitrary precision, which is crucial in real-world applications where the desired up-scaling factor may not be known beforehand.
Furthermore, Fig.~\ref{fig:ideal case on CelebAHQ} presents a visual comparison using RDN as the encoder. As the up-sampling scale increases, we find that INR-based SISR methods produce SR results with blocking artifacts when zooming into the SR images. In contrast, ARASFSR generates stable results with plausible details.
\noindent\textbf{Real-world case on CelebAHQ-NN-JPEG:} 
To evaluate the effectiveness in real-world scenarios, we assess the performance of INR-based SISR methods on CelebAHQ-NN-JPEG, synthesized by performing nearest neighbor down-sampling and augmenting with JPEG compression artifacts to mimic surveillance faces. We train the up-sampling scales within $\{\times1\sim\times2\}$ with a resolution setting of $L_{r}=32$ and $H_{r}$${\in}\{ 32\sim64\}$. To compare the performance of \emph{different input LR resolutions} during testing, we evaluate the models on the following $\times2$ super-resolution scenarios, using different input resolutions of 64(LR)-128(HR). Fig.~\ref{fig:real world case on CelebAHQ-NN-JPEG} offers a visual comparison with EDSR employed as the encoder. Previous works have encountered artifacts when the input resolution of facial images changes, as the implicit image function is constrained to a local view. The arrows indicate misplaced landmarks (such as smaller eyes) and artifacts (such as distortions in the nose). In contrast, our results display clear facial landmarks.

\begin{figure}[h!]
  \centering
  \includegraphics[width=\linewidth]{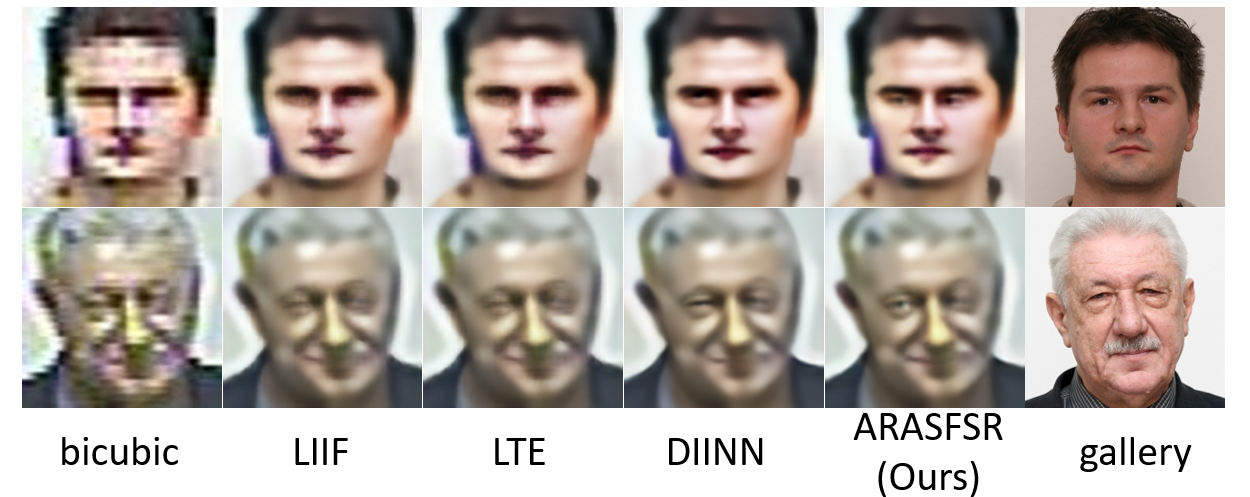}
  \caption{Visual comparison of real-world case on SCface with INR-based SISR methods. Please zoom in for details.}
  \label{fig:real world case on SCface}
\end{figure}

\noindent\textbf{Real-world case on SCface:} To further support the effectiveness in surveillance scenarios, we evaluate our approach using the d1 subset of SCface. This subset contains images captured by surveillance cameras from a distance of 4.2m, which is the most challenging subset in SCface. The visual comparison in Fig.~\ref{fig:real world case on SCface} shows the SR results of LR probes using different methods. Their HR gallery images are provided for comparison. It is evident from the comparison that existing methods encounter challenges in accurately preserving and reconstructing facial features, especially the eyes. In contrast, our method demonstrates exceptional performance in retaining and restoring fine facial details.

\subsubsection{Comparison with FSR methods}
We present a comparative analysis of our proposed method with SOTA FSR methods such as FSRNet \cite{CT-FSRNet-2018}, PFSRNet \cite{progressive-face-sr}, DICNet \cite{ma2020deep}, SPARNet \cite{ChenSPARNet}, SISN \cite{lu2021face} and MRRNet \cite{MRRNet}. 
In our comparison with FSR methods, we utilize RDN as a encoder. Since FSR methods focus on fixed-scale up-sampling, we train the networks for $\times8$ super-resolution with 
$L_{r}=16$ and $H_{r}=128$. However, during testing, we evaluate not only the match scenario (i.e., 16(LR)-128(HR)) but also the mismatch cases (i.e, 12(LR)-96(HR) and 64(LR)-128(HR)). As there is no standardized benchmark for comparing FSR methods, we take the following approach to ensure a fair comparison. For methods that provide training codes, we retrain these models using our training set on CelebAHQ and Helen. 
For methods that do not provide training codes, we test these models using their pretrained weights on CelebA in our CelebAHQ experiments. However, we do not evaluate the performance of these models on Helen since their training codes or pre-trained weights are not available.

\begin{table*}[h!]\small
\centering
\caption{Quantitative comparison on CelebAHQ and Helen with FSR methods. The best and second best performances are highlighted in \textcolor{red}{red} and \textcolor{blue}{blue} colors, respectively.}
\label{tab:ideal case on CelebAHQ and Helen with FSR}
\begin{tabular}{c|cccccc|ccc}
                                    & \multicolumn{6}{c|}{CelebAHQ}                                                                                                                                                        & \multicolumn{2}{c}{Helen}                                  &  \\ \cline{2-9}
                                    & \multicolumn{2}{c}{mismatch}                               & \multicolumn{2}{c}{match}                                  & \multicolumn{2}{c|}{mismatch}                              & \multicolumn{2}{c}{match}                                  &  \\
Method                              & \multicolumn{2}{c}{×8}                                     & \multicolumn{2}{c}{×8}                                     & \multicolumn{2}{c|}{×8}                                    & \multicolumn{2}{c}{×8}                                     &  \\
                                    & \multicolumn{2}{c}{12-96}                                  & \multicolumn{2}{c}{16-128}                                 & \multicolumn{2}{c|}{32-256}                                & \multicolumn{2}{c}{16-128}                                 &  \\ \cline{2-9}
                                    & \multicolumn{1}{c}{PSNR}                         & \multicolumn{1}{c}{SSIM}    & \multicolumn{1}{c}{PSNR}                         & \multicolumn{1}{c}{SSIM}    & \multicolumn{1}{c}{PSNR}                         & \multicolumn{1}{c|}{SSIM}   & \multicolumn{1}{c}{PSNR}                         & \multicolumn{1}{c}{SSIM}    &  \\ \hhline{=========}
FSRNet \cite{CT-FSRNet-2018}       & 22.4421                      & 0.6562                      & 24.4323                      & 0.7512                      & 25.1409                      & 0.7479                      & -                      & -                      &  \\
PFSRNet \cite{progressive-face-sr} & 22.6620                      & 0.6440                      & 20.0664                      & 0.5545                      & 22.2394                      & 0.6108                      & -                      & -                      &  \\
DICNet \cite{ma2020deep}           & 22.4446                      & 0.6112                      & 26.4781                      & 0.7758                      & 27.2324                      & 0.7631                      & 25.1722                      & 0.7265                      &  \\
SPARNet \cite{ChenSPARNet}         & 24.5753                      & 0.7086                      & 26.5321                      & 0.7792                      & 28.2184                      & 0.7907                      & \textcolor{blue}{25.5450}                      & \textcolor{red}{0.7509}                      &  \\
SISN \cite{lu2021face}             & 24.7083                      & 0.7166                      & \textcolor{red}{26.6575}  & \textcolor{blue}{0.7807} & \textcolor{blue}{28.3367} & \textcolor{blue}{0.7946} & 24.9867  & 0.7160 &  \\
MRRNet \cite{MRRNet}               & \textcolor{blue}{24.7207} & \textcolor{blue}{0.7192} & 26.5140                      & 0.7755                      & \textcolor{red}{28.3538}  & 0.7943                      & 25.1263                      & 0.7260                      &  \\
ARASFSR                    & \textcolor{red}{24.7712}  & \textcolor{red}{0.7214}  & \textcolor{blue}{26.6400} & \textcolor{red}{0.7813}  & 28.3088                      & \textcolor{red}{0.7948}  & \textcolor{red}{25.6002} & \textcolor{blue}{0.7445}  & 
\end{tabular}
\end{table*}

\begin{figure*}[h!]
  \centering
  \includegraphics[width=\linewidth]{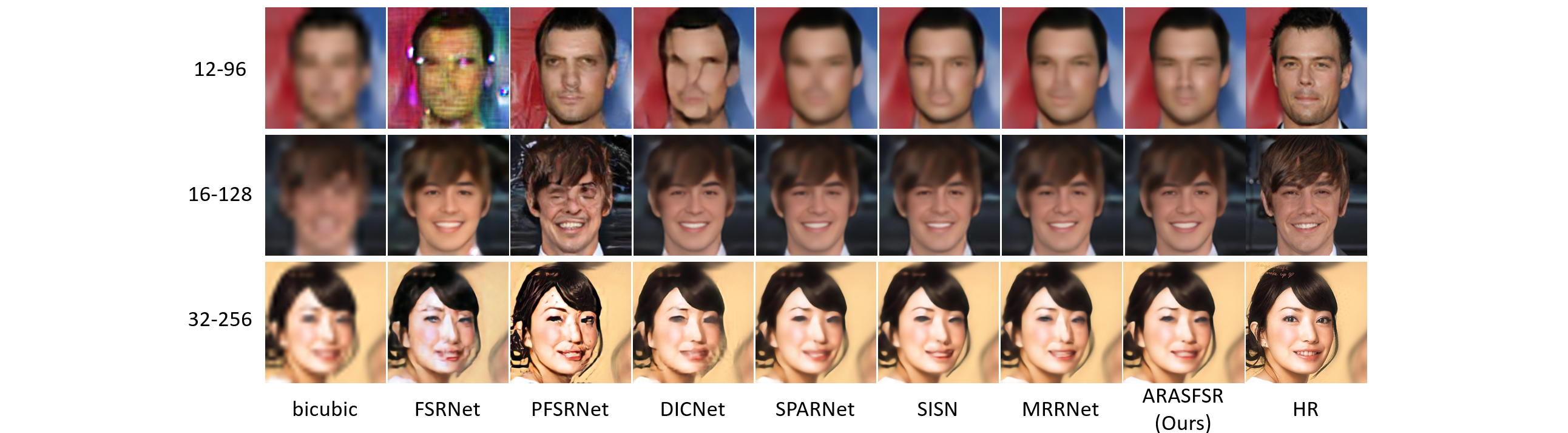}
  \caption{Visual comparison on CelebAHQ with FSR methods.}
  \label{fig:ideal case on CelebAHQ with FSR}
\end{figure*}

\noindent\textbf{Evaluation on CelebAHQ:} 
We first conduct experiments on CelebAHQ. The quantitative results presented in Table \ref{tab:ideal case on CelebAHQ and Helen with FSR} indicate that our method produces comparable performance to other FSR methods in both the match and mismatch scenarios. 
The visual comparison shown in Fig.~\ref{fig:ideal case on CelebAHQ with FSR} illustrates that our method produces high-quality results with minimal artifacts and clearer facial features. Unlike existing FSR methods that are only suitable for fixed up-sampling scales and input resolutions, our method can super-resolve images at any desired scale and is robust to input resolution, making it more versatile.

\noindent\textbf{Evaluation on Helen:} 
We conducted additional experiments on Helen in the match scenario (i.e., 16(LR)-128(HR)). The quantitative results in Table \ref{tab:ideal case on CelebAHQ and Helen with FSR} demonstrate that our method achieves results comparable to those of other FSR methods. Notably, our method can also be used with a stronger encoder to achieve even better results.

\begin{table}[h!]\small
\centering
\caption{Quantitative Ablation study of ARASFSR on CelebAHQ (PSNR(dB)).}
\label{tab:Ablation Study}
\begin{tabular}{c|cc|cc}
          & \multicolumn{2}{c|}{In-distribution} & \multicolumn{2}{c}{Out-of-distribution} \\
Method    & ×1.5          & ×2            & ×4        & ×8        \\
          & 64-96         & 64-128        & 64-256    & 64-512    \\ \hhline{=====}
ARASFSR     & \emph{40.1954}       & \emph{37.1097}       & \emph{32.4799}   & \emph{30.9148}   \\
ARASFSR(-L) & 40.1827       & 37.0936       & 32.3301   & 30.6935   \\
ARASFSR(-G) & 40.1666       & 37.0848       & 32.3210   & 30.6504   \\
ARASFSR(-S) & 40.1336       & 36.9969       & 32.3958   & 30.8488   
\end{tabular}
\end{table}

\begin{figure}[h!]
  \centering
  \includegraphics[width=\linewidth]{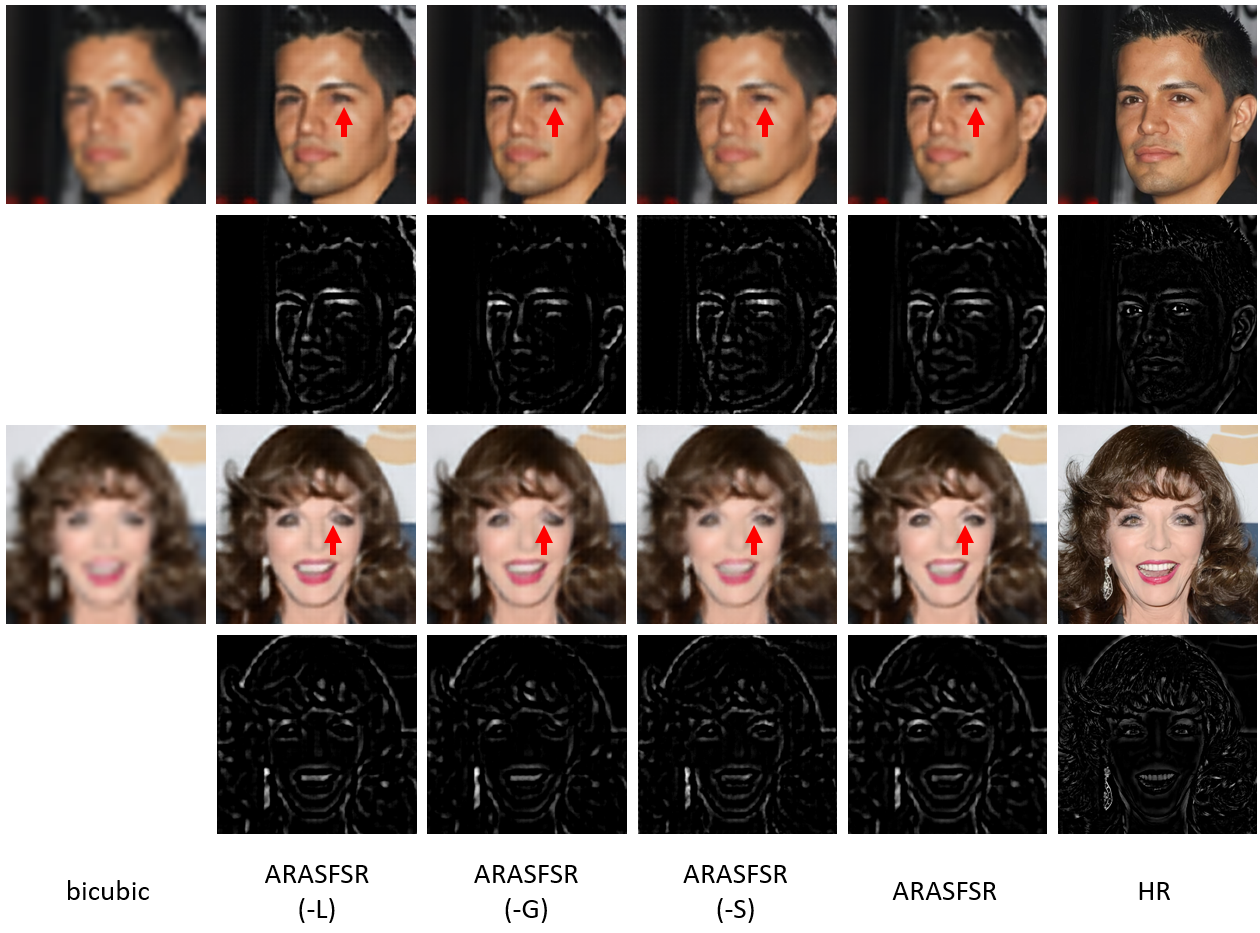}
  \caption{Qualitative ablation study of ARASFSR on CelebAHQ (LR-HR: 32-256). For a better view of the details, please zoom in.}
  \label{fig:ablation}
\end{figure}

\subsection{Ablation Study}
ARASFSR comprises implicit representation networks with three primary modules. We retrain the model with EDSR for three variants: ARASFSR without the local frequency estimation module (-L), ARASFSR without the global coordinate modulation module (-G), and ARASFSR without a skip connection (-S). To assess the effectiveness of each module, we conduct a quantitative ablation study, with the results presented in Table \ref{tab:Ablation Study}. Additionally, a qualitative ablation study depicted in Fig.~\ref{fig:ablation} examines the super-resolved results and the difference maps between the bicubic up-sampled results and the super-resolved results.

Our findings reveal that the local frequency estimation module plays a crucial role in capturing frequency information for facial texture, as evidenced by the comparison between ARASFSR and ARASFSR(-L). Moreover, the use of global coordinate modulation provides valuable guidance for fusing local features and global facial landmark priors, as demonstrated by the contrast between ARASFSR and ARASFSR(-G). Finally, we observe that the incorporation of a skip connection consistently enhances the quality of ARASFSR compared to ARASFSR(-S).

\section{Conclusion}
In this paper, we proposed an Arbitrary-Resolution and Arbitrary-Scale FSR method. Our framework employs an implicit image function that can effectively handle changes in the up-sampling scale. Furthermore, our approach incorporates a local frequency estimation module to capture high-frequency information for facial texture and a global coordinate modulation module to guide the facial structure, ensuring effective operation across various input sizes. Through quantitative and qualitative experiments, we demonstrated the robustness and superiority of our proposed method compared to state-of-the-art methods.

\noindent\textbf{Acknowledgements.}
Thanks to Yi-ren Ye and Hsuan-Tung Liu for valuable discussions of this work. This work was funded in part by E-SUN COMMERCIAL BANK, LTD. and also supported in part by the National Science and Technology Council, Taiwan, under Grant NSTC-112-2221-E-A49-089-MY3, Grant NSTC-110-2221-E-A49-066-MY3, Grant NSTC-111-2634-F-A49-010, Grant NSTC-112-2425-H-A49-001, and in part by the Higher Education Sprout Project of the National Yang Ming Chiao Tung University and the Ministry of Education (MOE), Taiwan.

\clearpage
{\small
\bibliographystyle{ieee_fullname}
\bibliography{egbib}

\begin{thebibliography}{10}\itemsep=-1pt

\bibitem{bao2022distilling}
Qiqi Bao, Rui Zhu, Bowen Gang, Pengyang Zhao, Wenming Yang, and Qingmin Liao.
\newblock Distilling resolution-robust identity knowledge for texture-enhanced face hallucination.
\newblock In {\em Proceedings of the 30th ACM International Conference on Multimedia}, pages 6727--6736, 2022.

\bibitem{Vggface2}
Qiong Cao, Li Shen, Weidi Xie, Omkar~M Parkhi, and Andrew Zisserman.
\newblock Vggface2: A dataset for recognising faces across pose and age.
\newblock In {\em 2018 13th IEEE international conference on automatic face \& gesture recognition (FG 2018)}, pages 67--74. IEEE, 2018.

\bibitem{ChenSPARNet}
Chaofeng Chen, Dihong Gong, Hao Wang, Zhifeng Li, and Kwan-Yee~K. Wong.
\newblock Learning spatial attention for face super-resolution.
\newblock 2020.

\bibitem{IdentityAware}
Jin Chen, Jun Chen, Zheng Wang, Chao Liang, and Chia-Wen Lin.
\newblock Identity-aware face super-resolution for low-resolution face recognition.
\newblock {\em IEEE Signal Processing Letters}, 27:645--649, 2020.

\bibitem{DBLP:journals/corr/abs-2012-09161}
Yinbo Chen, Sifei Liu, and Xiaolong Wang.
\newblock Learning continuous image representation with local implicit image function.
\newblock In {\em Proceedings of the IEEE/CVF Conference on Computer Vision and Pattern Recognition}, pages 8628--8638, 2021.

\bibitem{CT-FSRNet-2018}
Yu* Chen, Ying* Tai, Xiaoming Liu, Chunhua Shen, and Jian Yang.
\newblock Fsrnet: End-to-end learning face super-resolution with facial priors.
\newblock In {\em Proceedings of the IEEE Conference on Computer Vision and Pattern Recognition}, 2018.

\bibitem{chen2018implicit_decoder}
Zhiqin Chen and Hao Zhang.
\newblock Learning implicit fields for generative shape modeling.
\newblock {\em Proceedings of IEEE Conference on Computer Vision and Pattern Recognition (CVPR)}, 2019.

\bibitem{chibane20ifnet}
Julian Chibane, Thiemo Alldieck, and Gerard Pons-Moll.
\newblock Implicit functions in feature space for 3d shape reconstruction and completion.
\newblock In {\em {IEEE} Conference on Computer Vision and Pattern Recognition (CVPR)}. {IEEE}, jun 2020.

\bibitem{progressive-face-sr}
Kim Deokyun, Kim Minseon, Kwon Gihyun, and Kim Dae-Shik.
\newblock Progressive face super-resolution via attention to facial landmark.
\newblock In {\em Proceedings of the 30th British Machine Vision Conference (BMVC)}, 2019.

\bibitem{Genova_2020_CVPR}
Kyle Genova, Forrester Cole, Avneesh Sud, Aaron Sarna, and Thomas Funkhouser.
\newblock Local deep implicit functions for 3d shape.
\newblock In {\em IEEE/CVF Conference on Computer Vision and Pattern Recognition (CVPR)}, June 2020.

\bibitem{grgic2011scface}
Mislav Grgic, Kresimir Delac, and Sonja Grgic.
\newblock Scface--surveillance cameras face database.
\newblock {\em Multimedia tools and applications}, 51(3):863--879, 2011.

\bibitem{huang2017wavelet}
Huaibo Huang, Ran He, Zhenan Sun, and Tieniu Tan.
\newblock Wavelet-srnet: A wavelet-based cnn for multi-scale face super resolution.
\newblock In {\em Proceedings of the IEEE international conference on computer vision}, pages 1689--1697, 2017.

\bibitem{MRRNet}
Weikang Huang, Shiyong Lan, Wenwu Wang, Xuedong Yuan, Hongyu Yang, Piaoyang Li, and Wei Ma.
\newblock Face super-resolution with spatial attention guided by multiscale receptive-field features.
\newblock In {\em International Conference on Artificial Neural Networks,(ICANN2022)}, 2022.

\bibitem{Local_Implicit_Grid_CVPR20}
Chiyu~Max Jiang, Avneesh Sud, Ameesh Makadia, Jingwei Huang, Matthias Nießner, and Thomas Funkhouser.
\newblock Local implicit grid representations for 3d scenes.
\newblock In {\em Proceedings IEEE Conf. on Computer Vision and Pattern Recognition (CVPR)}, 2020.

\bibitem{karras2017progressive}
Tero Karras, Timo Aila, Samuli Laine, and Jaakko Lehtinen.
\newblock Progressive growing of gans for improved quality, stability, and variation.
\newblock {\em arXiv preprint arXiv:1710.10196}, 2017.

\bibitem{kim2016accurate}
Jiwon Kim, Jung~Kwon Lee, and Kyoung~Mu Lee.
\newblock Accurate image super-resolution using very deep convolutional networks.
\newblock In {\em Proceedings of the IEEE conference on computer vision and pattern recognition}, pages 1646--1654, 2016.

\bibitem{kingma2014adam}
Diederik~P Kingma and Jimmy Ba.
\newblock Adam: A method for stochastic optimization.
\newblock {\em arXiv preprint arXiv:1412.6980}, 2014.

\bibitem{IdentityPreserved}
Shun-Cheung Lai, Chen-Hang He, and Kin-Man Lam.
\newblock Low-resolution face recognition based on identity-preserved face hallucination.
\newblock In {\em 2019 IEEE International Conference on Image Processing (ICIP)}, pages 1173--1177. IEEE, 2019.

\bibitem{lai2017deep}
Wei-Sheng Lai, Jia-Bin Huang, Narendra Ahuja, and Ming-Hsuan Yang.
\newblock Deep laplacian pyramid networks for fast and accurate super-resolution.
\newblock In {\em Proceedings of the IEEE conference on computer vision and pattern recognition}, pages 624--632, 2017.

\bibitem{le2012interactive}
Vuong Le, Jonathan Brandt, Zhe Lin, Lubomir Bourdev, and Thomas~S Huang.
\newblock Interactive facial feature localization.
\newblock In {\em Computer Vision--ECCV 2012: 12th European Conference on Computer Vision, Florence, Italy, October 7-13, 2012, Proceedings, Part III 12}, pages 679--692. Springer, 2012.

\bibitem{lte-jaewon-lee}
Jaewon Lee and Kyong~Hwan Jin.
\newblock Local texture estimator for implicit representation function.
\newblock In {\em Proceedings of the IEEE/CVF Conference on Computer Vision and Pattern Recognition (CVPR)}, pages 1929--1938, June 2022.

\bibitem{lim2017enhanced}
Bee Lim, Sanghyun Son, Heewon Kim, Seungjun Nah, and Kyoung Mu~Lee.
\newblock Enhanced deep residual networks for single image super-resolution.
\newblock In {\em Proceedings of the IEEE conference on computer vision and pattern recognition workshops}, pages 136--144, 2017.

\bibitem{liu2015deep}
Ziwei Liu, Ping Luo, Xiaogang Wang, and Xiaoou Tang.
\newblock Deep learning face attributes in the wild.
\newblock In {\em Proceedings of the IEEE international conference on computer vision}, pages 3730--3738, 2015.

\bibitem{lu2021face}
Tao Lu, Yuanzhi Wang, Yanduo Zhang, Yu Wang, Liu Wei, Zhongyuan Wang, and Junjun Jiang.
\newblock Face hallucination via split-attention in split-attention network.
\newblock In {\em Proceedings of the 29th ACM International Conference on Multimedia}, pages 5501--5509, 2021.

\bibitem{ma2020deep}
Cheng Ma, Zhenyu Jiang, Yongming Rao, Jiwen Lu, and Jie Zhou.
\newblock Deep face super-resolution with iterative collaboration between attentive recovery and landmark estimation.
\newblock In {\em Proceedings of the IEEE Conference on Computer Vision and Pattern Recognition (CVPR)}, 2020.

\bibitem{9776607}
Cheng Ma, Peiqi Yu, Jiwen Lu, and Jie Zhou.
\newblock Recovering realistic details for magnification-arbitrary image super-resolution.
\newblock {\em IEEE Transactions on Image Processing}, 31:3669--3683, 2022.

\bibitem{mehta2021modulated}
Ishit Mehta, Micha{\"e}l Gharbi, Connelly Barnes, Eli Shechtman, Ravi Ramamoorthi, and Manmohan Chandraker.
\newblock Modulated periodic activations for generalizable local functional representations.
\newblock In {\em Proceedings of the IEEE/CVF International Conference on Computer Vision}, pages 14214--14223, 2021.

\bibitem{mildenhall2020nerf}
Ben Mildenhall, Pratul~P. Srinivasan, Matthew Tancik, Jonathan~T. Barron, Ravi Ramamoorthi, and Ren Ng.
\newblock Nerf: Representing scenes as neural radiance fields for view synthesis.
\newblock In {\em ECCV}, 2020.

\bibitem{nguyen2023single}
Quan~H. Nguyen and William~J Beksi.
\newblock Single image super-resolution via a dual interactive implicit neural network.
\newblock In {\em Proceedings of the IEEE/CVF Winter Conference on Applications of Computer Vision (WACV)}, pages 4936--4945, 2023.

\bibitem{rahaman2019spectral}
Nasim Rahaman, Aristide Baratin, Devansh Arpit, Felix Draxler, Min Lin, Fred Hamprecht, Yoshua Bengio, and Aaron Courville.
\newblock On the spectral bias of neural networks.
\newblock In {\em International Conference on Machine Learning}, pages 5301--5310. PMLR, 2019.

\bibitem{sitzmann2020implicit}
Vincent Sitzmann, Julien Martel, Alexander Bergman, David Lindell, and Gordon Wetzstein.
\newblock Implicit neural representations with periodic activation functions.
\newblock {\em Advances in Neural Information Processing Systems}, 33:7462--7473, 2020.

\bibitem{sitzmann2019scene}
Vincent Sitzmann, Michael Zollh{\"o}fer, and Gordon Wetzstein.
\newblock Scene representation networks: Continuous 3d-structure-aware neural scene representations.
\newblock {\em Advances in Neural Information Processing Systems}, 32, 2019.

\bibitem{wang2004image}
Zhou Wang, Alan~C Bovik, Hamid~R Sheikh, and Eero~P Simoncelli.
\newblock Image quality assessment: from error visibility to structural similarity.
\newblock {\em IEEE transactions on image processing}, 13(4):600--612, 2004.

\bibitem{xu2021ultrasr}
Xingqian Xu, Zhangyang Wang, and Humphrey Shi.
\newblock Ultrasr: Spatial encoding is a missing key for implicit image function-based arbitrary-scale super-resolution.
\newblock {\em arXiv preprint arXiv:2103.12716}, 2021.

\bibitem{yu2016ultra}
Xin Yu and Fatih Porikli.
\newblock Ultra-resolving face images by discriminative generative networks.
\newblock In {\em Computer Vision--ECCV 2016: 14th European Conference, Amsterdam, The Netherlands, October 11-14, 2016, Proceedings, Part V}, pages 318--333. Springer, 2016.

\bibitem{superIdentity}
Kaipeng Zhang, Zhanpeng Zhang, Chia-Wen Cheng, Winston~H Hsu, Yu Qiao, Wei Liu, and Tong Zhang.
\newblock Super-identity convolutional neural network for face hallucination.
\newblock In {\em Proceedings of the European conference on computer vision (ECCV)}, pages 183--198, 2018.

\bibitem{7553523}
Kaipeng Zhang, Zhanpeng Zhang, Zhifeng Li, and Yu Qiao.
\newblock Joint face detection and alignment using multitask cascaded convolutional networks.
\newblock {\em IEEE signal processing letters}, 23(10):1499--1503, 2016.

\bibitem{zhang2018residual}
Yulun Zhang, Yapeng Tian, Yu Kong, Bineng Zhong, and Yun Fu.
\newblock Residual dense network for image super-resolution.
\newblock In {\em Proceedings of the IEEE conference on computer vision and pattern recognition}, pages 2472--2481, 2018.

\end{thebibliography}
}

\end{document}